\documentclass[10pt,twocolumn,letterpaper]{article}

\usepackage{cvpr}              

\usepackage{graphicx}
\usepackage{amsmath}
\usepackage{amssymb}
\usepackage{booktabs}
\usepackage{multirow}
\usepackage{bbding}
\usepackage{verbatim}
\usepackage{floatrow}
\usepackage[accsupp]{axessibility}
\newfloatcommand{capbtabbox}{table}[][\FBwidth]

\usepackage[pagebackref,breaklinks,colorlinks]{hyperref}

\usepackage[capitalize]{cleveref}
\crefname{section}{Sec.}{Secs.}
\Crefname{section}{Section}{Sections}
\Crefname{table}{Table}{Tables}
\crefname{table}{Tab.}{Tabs.}

\def\cvprPaperID{6796} 
\def\confName{CVPR}
\def\confYear{2022}

\begin{document}

\title{Rethinking the Augmentation Module in Contrastive Learning: Learning Hierarchical Augmentation Invariance with Expanded Views}

\author{Junbo Zhang~~~~~Kaisheng Ma\thanks{Corresponding Author.}\\
Tsinghua University\\
{\tt\small zhangjb21@mails.tsinghua.edu.cn}~~~~{\tt\small kaisheng@mail.tsinghua.edu.cn}
}
\maketitle

\begin{abstract}
A data augmentation module is utilized in contrastive learning to transform the given data example into two views, which is considered essential and irreplaceable. However, the pre-determined composition of multiple data augmentations brings two drawbacks. First, the artificial choice of augmentation types brings specific representational invariances to the model, which have different degrees of positive and negative effects on different downstream tasks. Treating each type of augmentation equally during training makes the model learn non-optimal representations for various downstream tasks and limits the flexibility to choose augmentation types beforehand. Second, the strong data augmentations used in classic contrastive learning methods may bring too much invariance in some cases, and fine-grained information that is essential to some downstream tasks may be lost. This paper proposes a general method to alleviate these two problems by considering “where” and “what” to contrast in a general contrastive learning framework. We first propose to learn different augmentation invariances at different depths of the model according to the importance of each data augmentation instead of learning representational invariances evenly in the backbone. We then propose to expand the contrast content with augmentation embeddings to reduce the misleading effects of strong data augmentations. Experiments based on several baseline methods demonstrate that we learn better representations for various benchmarks on classification, detection, and segmentation downstream tasks.
\end{abstract}

\section{Introduction}
\label{Introduction}
Contrastive learning has been proved to be able to learn meaningful visual representations without human annotations~\cite{Chen2021ExploringSS,Grill2020BootstrapYO}. Original methods regard two views transformed from the same example as a positive pair and other examples in the batch or the memory bank as negative samples~\cite{Oord2018RepresentationLW}, then the model is trained with the contrastive loss. Many techniques were previously considered important in this process, such as strong data augmentations~\cite{Chen2020ASF}, the selection of negative samples~\cite{Hu2020AdCoAC,Wu2020OnMI}, momentum-update encoder~\cite{He2020MomentumCF,Grill2020BootstrapYO}, and training details like large batch sizes and long training epochs~\cite{Chen2020ASF}. However, recent works prove that useful visual representations can be learned without negative pairs, the momentum-update of parameters or large batch size ~\cite{Grill2020BootstrapYO,Chen2021ExploringSS,Zbontar2021BarlowTS}. The most indispensable process in contrastive learning is the data augmentation module. The essential principle of contrastive learning is to learn the representational invariance by making the network learn to be invariant to a set of data augmentations~\cite{Tian2020WhatMF}.

\begin{figure}[t]
\vskip 0.1in
\begin{center}
\centerline{\includegraphics[width=0.87\columnwidth]{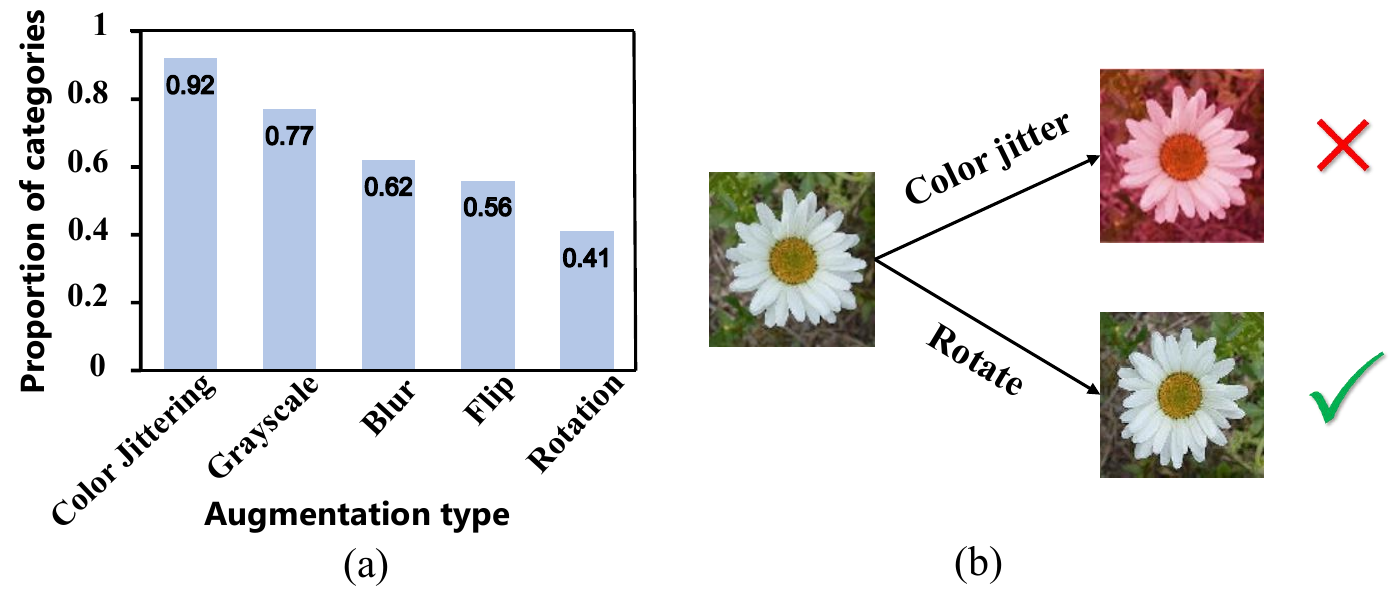}}
\caption{The influence of augmentation types. (a) The proportion of categories in ImageNet where the data augmentation has a positive effect. (b) An example of the flower where color invariance has a negative effect and rotation invariance has a positive effect. The conclusion is reversed for most categories in ImageNet.}
\label{fig1}
\end{center}
\vspace{-0.8cm}
\end{figure}

Previous works show that the composition of multiple types of data augmentations is crucial for contrastive learning~\cite{Chen2020ASF}. A specific set of augmentations is determined after extensive experiments to achieve the best results on large-scale datasets (e.g., ImageNet). In most recent works~\cite{He2020MomentumCF,Sordoni2021DecomposedMI,Wang2021DenseCL,Jiang2021SelfDamagingCL,Bukchin2021FinegrainedAC}, the data augmentation pipeline consists of random cropping and resizing, horizontal flipping, color jittering, converting to grayscale, and Gaussian blurring. However, the pre-determined and artificial choices of augmentation types and augmentation strength bring the corresponding problems as follows.

\begin{figure}[t]
\vskip 0.1in
\begin{center}
\centerline{\includegraphics[width=0.9\columnwidth]{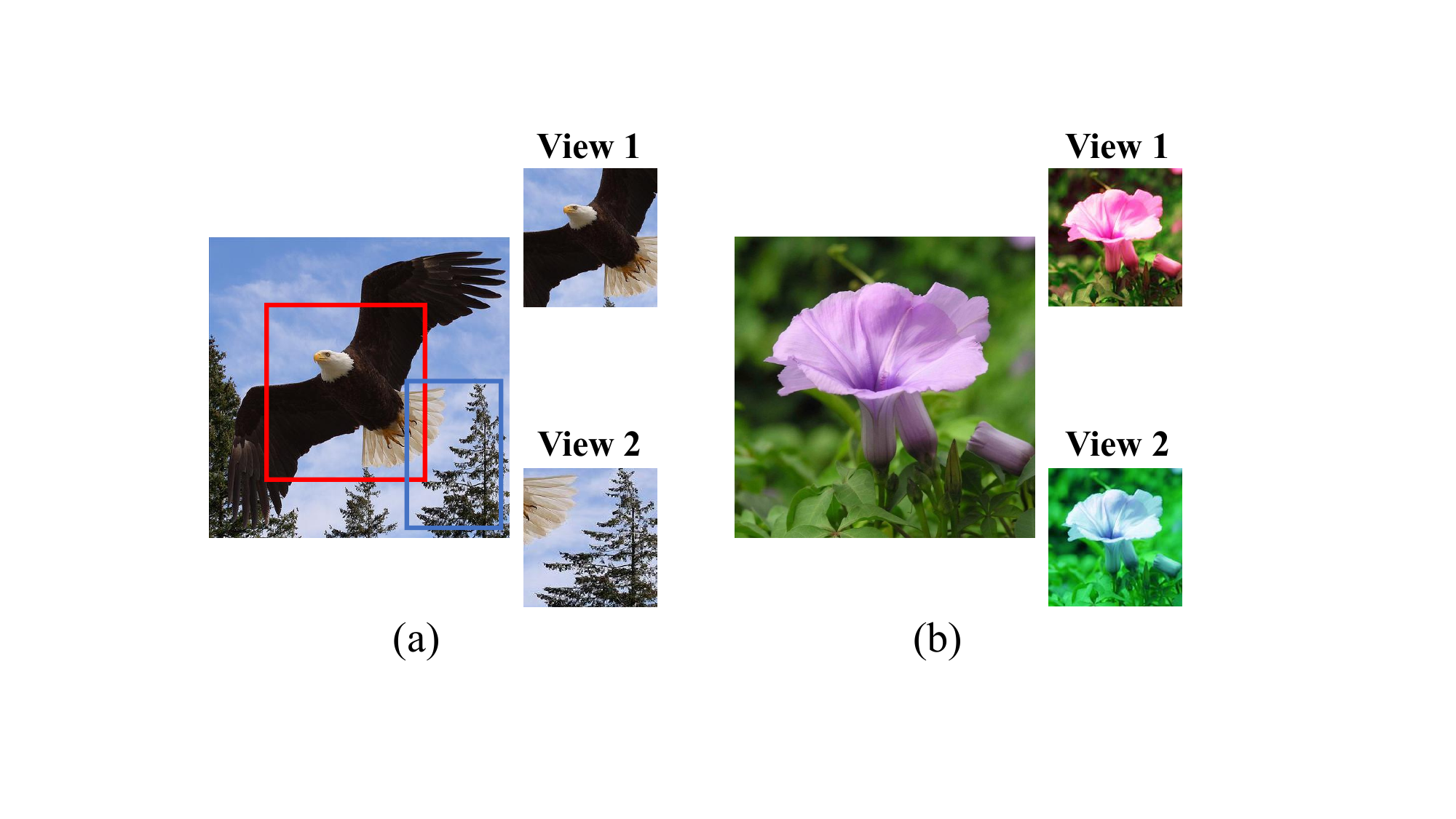}}
\caption{The influence of augmentation strength. (a) Aligning the two cropped views introduces misleading spatial information. (b) Aligning the two color-augmented views loses the fine-grained color information of flowers.}
\label{fig2}
\end{center}
\vspace{-0.8cm}
\vskip -0.05in
\end{figure}

In terms of the augmentation type, after selecting specific types, current contrastive learning methods treat each augmentation equally, and all the representational invariances are evenly distributed in the backbone. However, according to previous works, some augmentations are fundamental for contrastive learning~\cite{Chen2020ASF,Grill2020BootstrapYO}, such as cropping and color jittering, while others are less important or even harmful, such as blurring and rotating. Figure~\ref{fig1} (a) shows the proportion of categories in ImageNet where the data augmentation has a positive effect. Furthermore, each data augmentation may have a different impact on downstream tasks. As shown in Figure~\ref{fig1} (b), using color jittering may help with most categories in ImageNet but harms the representation learning of flowers, while adding rotation brings the opposite effect. Thus, simply getting rid of one augmentation (e.g., rotation) and keeping the other as in the classical augmentation pipeline assumes an implicit knowledge of invariance for no reason. Indiscriminately learning the augmentation invariances in the encoder makes it less flexible in choosing augmentation types beforehand and may lead to non-optimal representations for various downstream tasks.

In terms of the augmentation strength, SimCLR~\cite{Chen2020ASF} shows that contrastive learning benefits from strong data augmentations. InfoMin principle~\cite{Tian2020WhatMF} also argues that weak augmentations bring too much noise into the mutual information between two augmented views, which leads to worse generalization on the downstream task. Nevertheless, the representations learned using strong augmentations may sometimes bring unnecessary invariance into the backbone and lose fine-grained information essential to some downstream tasks. As shown in Figure~\ref{fig2}, projecting the two cropped views into the same position in the feature space learns misleading spatial information and brings too much invariance. And projecting the two color-augmented views into the same position loses the fine-grained category information~\cite{Ericsson2021HowWD} (e.g., the color of morning glory). Although classical methods implicitly alleviate this problem by adding a projection head after the encoder, no explicit method has been proposed in unsupervised contrastive learning.

This paper proposes a generic method to tackle these two augmentation-related problems by considering “where” and “what” to contrast. First, we propose to treat various data augmentations differently and learn the “hierarchical augmentation invariance”. By computing multiple contrastive losses at different depths of the encoder, we make the fundamental augmentation invariances more widely distributed and some generally insignificant invariances restricted to the deeper layers. By restricting the impact scope of data augmentation without weakening its strength, we demonstrate that adding a specific type of data augmentation that was not used in the classical augmentation set can simultaneously improve the performance on both large-scale datasets (e.g., ImageNet, COCO) and fine-grained datasets (e.g., VGG Flowers, iNaturelist-2019). Second, we propose to expand the contrast content with augmentation embeddings. By augmenting the original labels with input transformation, the label augmentation~\cite{Lee2020SelfsupervisedLA} method in supervised learning relaxes specific transformation invariant constraints and prevents the loss of transformation-related information. Inspired by this, we regard each view as the “label” of the other view and expand the extracted view features with corresponding augmentation embeddings. The encoded augmentation information helps to reduce the unnecessary invariance and make up for some lost fine-grained information. The small network for embedding augmentation parameters is simultaneously trained with the encoder and is discarded during the inference. Our analysis demonstrates that the augmentation embeddings learn useful information of specific augmentations and benefit the representation learning in various benchmarks.

We apply our method to several baseline contrastive learning architectures and evaluate the representations on various classification, detection, and segmentation benchmarks. Our results reveal that the proposed method consistently improves the performance compared to baselines on various downstream tasks.

\section{Related work}
\label{sec:Related}

\subsection{Contrastive learning}
Contrastive learning aims to learn generalizable and transferable representations from unlabeled data using contrastive pairs. The classic method constructs a positive pair and a negative pair for each data sample and optimizes the model with infoNCE loss~\cite{Oord2018RepresentationLW}. The choice of negative samples is considered important in this process. MoCo~\cite{He2020MomentumCF} introduces a dynamic memory bank to record the embeddings of negative samples. AdCo~\cite{Hu2020AdCoAC} learns a set of negative samples by adversarial training. Other works propose to contrast the data with prototypical representations trying to find more suitable negative samples for contrastive learning~\cite{Caron2020UnsupervisedLO,Li2021PrototypicalCL}. Besides negative samples, SimCLR~\cite{Chen2020ASF} shows the large batch size and long training time are also crucial for contrastive learning. Beyond the aforementioned classic training framework, BYOL~\cite{Grill2020BootstrapYO} introduces a slow-moving average network and shows that contrastive learning can be effective without using any negative samples. By introducing stop-gradient, SimSiam~\cite{Chen2021ExploringSS} shows that simple Siamese networks can learn meaningful representations even without negative pairs, large batches, or momentum encoders. Barlow Twins~\cite{Zbontar2021BarlowTS} further proposes a new contrastive learning objective without using stop-gradient, which also brings comparable results. Recently, some theoretical works have tried to understand how these new methods succeed in avoiding representational collapse~\cite{Tian2021UnderstandingSL,Hua2021OnFD}. Although contrastive learning framework has been dramatically simplified, producing a contrastive pair with multiple types of data augmentations is still considered vital and indispensable in most works. The essence of contrastive learning is to learn multiple augmentation invariances in the representations so that these representations can be utilized successfully in a variety of downstream tasks.

\subsection{Augmentation in contrastive learning}
The augmentation module in contrastive learning transforms the given data sample into two correlated views. SimCLR~\cite{Chen2020ASF} first shows that the composition of multiple data augmentations is crucial to yield effective representations, among which the composition of random cropping and color jittering stands out. The paper also conducts detailed experiments to confirm that unsupervised contrastive learning benefits from strong data augmentation and determines a certain set of augmentations that yield the best results on universal datasets. BYOL~\cite{Grill2020BootstrapYO} proposes a novel architecture that is more robust to the choice of image augmentations. Its performance is much less affected than SimCLR when removing some augmentations. However, simply removing one type of augmentation in BYOL can still result in a 5\% $\sim$ 25\% decrease in accuracy. As the necessity of data augmentation is fully confirmed, many works explore what kind of data augmentation can lead to better representations in contrastive learning. SwAV~\cite{Caron2020UnsupervisedLO} proposes a new ‘multi-crop’ augmentation strategy that mixes the views of different resolutions. CsMl~\cite{seeds} applies CutMix~\cite{Yun2019CutMixRS} augmentation to generate the views with cross-samples and multi-level representation. Besides, InfoMin~\cite{Tian2020WhatMF} argues that the optimal augmentation strategy should reduce the mutual information between views while maintaining the task-relevant information. Since this ‘optimal’ augmentation strategy is strongly related to downstream tasks, InfoMin proposes to learn this optimal strategy in a semi-supervised way. Although this method does not work well in a complete unsupervised framework, it inspires us to think of whether using a pre-determined composition of data augmentations, as in most previous works, has some drawbacks on certain downstream tasks. A similar idea is shown in MaskCo~\cite{Zhao2021SelfSupervisedVR}, which argues that the implicit semantic consistency assumption in instance discrimination pre-training task may harm the performance of downstream tasks on unconstrained datasets. In terms of augmentations, the pre-determined data augmentations assume particular representation invariances that may not always be needed in downstream tasks. LooC~\cite{Xiao2021WhatSN} weakens this assumption implicitly by using a multi-head network. Nevertheless, the backbone of LOOC still has the same representational invariances. Instead, we reduce the drawbacks of both the pre-determined augmentation types and augmentation strength by considering “where” and “what” to contrast in contrastive learning.

\section{Method}

\begin{figure*}[t]
\vskip 0.1in
\begin{center}
\centerline{\includegraphics[width=0.88\columnwidth]{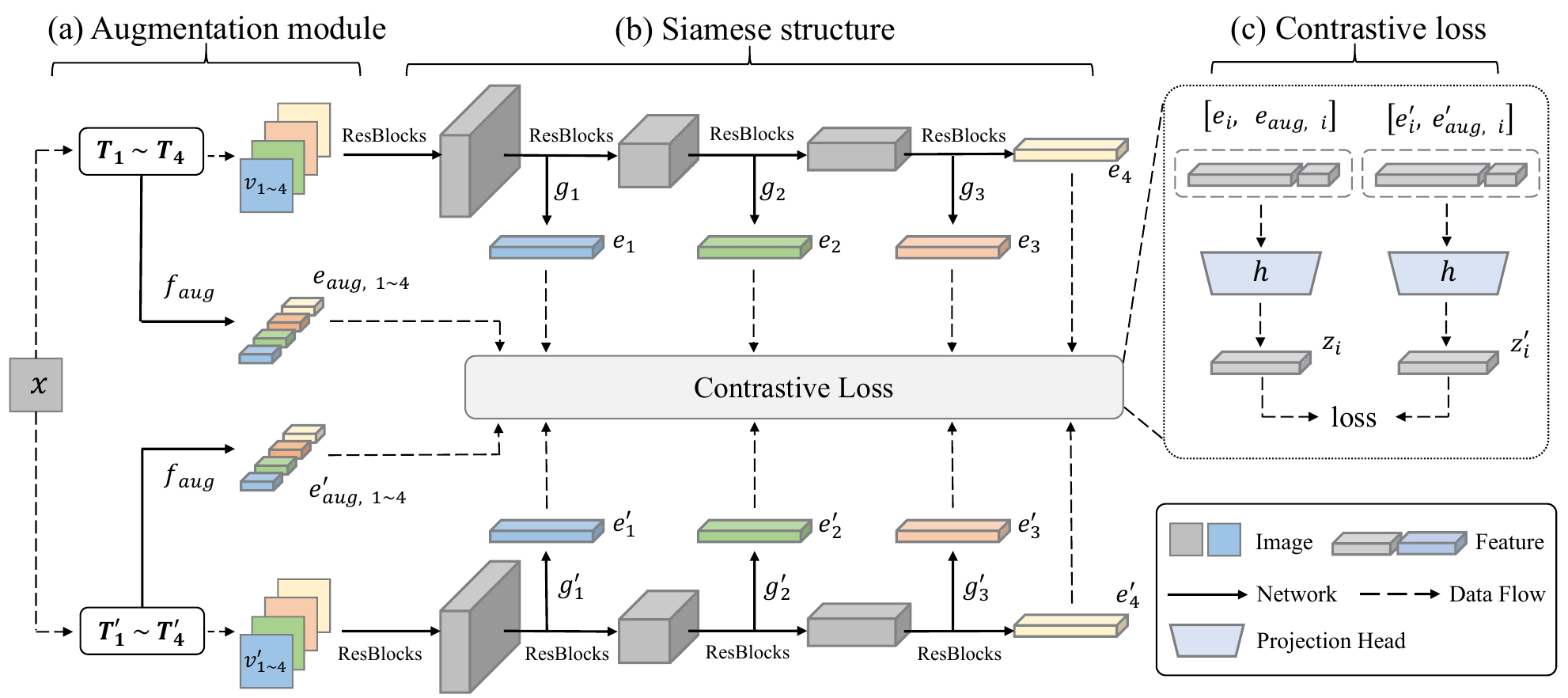}}
\caption{The improvements to the three components in the general contrastive learning framework. (a) Augmentation module: We generate multiple pairs of views with an add-one strategy and utilize a small network to embed the augmentation parameters. (b) Siamese structure: We divide the backbone into several stages according to the depth. Multiple pairs of view features shown in different colors are extracted at different stages. (c) Contrastive loss: We introduce multiple contrastive losses at different stages using multiple pairs of view features. The view features are expanded with corresponding augmentation embeddings before the projection head. The expanded features can then be used to calculate contrastive loss as in many previous works.}
\label{main_fig}
\end{center}
\vspace{-1cm}
\end{figure*}
\setlength{\belowcaptionskip}{-3cm} 

We first review the three indispensable components in the general contrastive learning framework in Section~\ref{general}: the augmentation module, the siamese structure, and the contrastive loss. Then we illustrate the overall architecture in Section~\ref{overeview} and show that the proposed method can be combined with any approach that fits within the general contrastive learning framework. Next, we thoroughly introduce the proposed methods dealing with the problems of augmentation types and augmentation strength in Section~\ref{hier} and~\ref{expansion}, respectively.

\subsection{General contrastive learning framework}
\label{general}
We first briefly review the three components in the general framework. Firstly, a data augmentation module transforms the given data example into two views randomly. These two correlated views are then fed to the backbone and are used to compute the contrastive loss. The composition of multiple data augmentations widely used in most previous works contains random cropping and resizing, horizontal flipping, color jittering, converting to grayscale, and Gaussian blurring. Secondly, two neural-network-based encoders form a siamese structure that extracts representation vectors from the two views. Note that the two encoders can share their weights~\cite{Chen2021ExploringSS,Zbontar2021BarlowTS} or can be two separate networks with the same structure~\cite{He2020MomentumCF,Verma2021TowardsDC,Misra2020SelfSupervisedLO}. The encoders can be optimized by backpropagation or the momentum-based moving average of parameters~\cite{Grill2020BootstrapYO}. Thirdly, a contrastive loss is used to define a contrastive prediction task. Several contrastive losses have been proposed in previous works, such as InfoNCE~\cite{Oord2018RepresentationLW} computed with the negative samples or prototypes~\cite{Caron2020UnsupervisedLO,Li2021PrototypicalCL}, negative cosine similarity between two positive view features~\cite{Chen2021ExploringSS}, and the similarity between the cross-correlation matrix of views and the identity matrix~\cite{Zbontar2021BarlowTS}. The core of these objective functions is to narrow the distance between two views in the feature space.

\subsection{Overall architecture of our method}
\label{overeview}
\vspace{-1.5mm}
The overview of our method is shown in Figure~\ref{main_fig}. We introduce three improvements to the contrastive learning framework. First, instead of transforming the data sample into one pair of views, we produce multiple augmentation modules with add-one strategy and transform the image into multiple pairs of views as in Figure~\ref{main_fig} (a). “Add-one” means each augmentation module has one more augmentation type than previous one. A small network is used to embed specific augmentation parameters for subsequent improvements. Second, the siamese structure is divided into several stages according to the depth as in Figure~\ref{main_fig} (b). The multiple pairs of view features will subsequently be extracted at different stages and used to compute multiple contrastive losses, which “supervise” different parts of the encoder. With these two improvements, hierarchical augmentation invariance can be learned in the encoder. Third, we expand the view features with augmentation embeddings while keeping the original loss function unchanged as in Figure~\ref{main_fig} (c). By doing so, the view information is combined with the augmentation information. Note that these three improvements can be combined with all variants of the current contrastive learning framework. We next introduce the first two improvements in Section~\ref{hier} and the third improvement in Section~\ref{expansion}.

\subsection{Hierarchical augmentation invariance}
\label{hier}
\vspace{-1.5mm}
In order to alleviate the problem of augmentation types as discussed in Section~\ref{Introduction}, we propose to learn the hierarchical augmentation invariances by applying different invariances to different depths of the encoder. Specifically, we take an add-one strategy based on the importance of each augmentation type to produce multiple augmentation modules. Thus, each augmentation module has one more augmentation type than the previous module, and the last module contains all the types of augmentations as in the classical pipeline. In this paper, we divide the backbone into four stages according to the depth as previous works~\cite{Zhang2019BeYO,Zhang2020TaskOrientedFD}. We formulate the complete augmentation pipeline in most previous works as $T = {\rm Compose}\{t_{0}, t_{1}, t_{2}, t_{3}, t_{4}\}$, where $t_{0}$ contains the base augmentations and $t_{1} \text{ } \sim t_{4}$ each represents one type of augmentation. So the four augmentation modules $T_{1}, T_{2}, T_{3}, T_{4}$ generated by add-one strategy can be formulated as:

{\setlength\abovedisplayskip{0cm}
\setlength\belowdisplayskip{0cm}
\begin{equation}
\label{transform}
\begin{split}
&T_{1} = {\rm Compose}\{t_{0},t_{1}\}, \\
&T_{2} = {\rm Compose}\{t_{0},t_{1},t_{2}\}, \\
&T_{3} = {\rm Compose}\{t_{0},t_{1},t_{2},t_{3}\}, \\ 
&T_{4} = {\rm Compose}\{t_{0},t_{1},t_{2},t_{3},t_{4}\}. 
\end{split}
\end{equation} }

With these modules, we augment the given data sample $x$ into eight views as: $v_{i} = T_{i}\left(x\right),\text{ } v_{i}^{’}=T_{i}^{’}\left(x\right), \text{ } i=1,2,3,4$, where $T_{i}$ and $T_{i}^{'}$ have the same augmentation types but are two random instances. In this way, we obtain four pairs of views. Aligning each pair of view features brings a specific composition of augmentation invariance. 

With these pairs of views, we propose to compute multiple contrastive losses at different depths of the encoder. Optimizing each loss function brings specific invariance to a subpart of the encoder. We introduce the contrastive losses at the end of each stage for the ResNet backbone. Specifically, we add several convolutional layers to the shallow stages to get features of the same shape as the last stage output. These extra layers also help to solve gradient competition issue~\cite{2015FitNets} and extract more meaningful visual representations. Then the four pairs of view features are extracted by the four subparts of the encoder respectively, which can be formulated as:

{\setlength\abovedisplayskip{0cm}
\setlength\belowdisplayskip{0cm}
\begin{equation}
\label{layer}
e_{i} = g_{i}(f_{i}(v_{i})); \text{ } e_{i}^{’} = g_{i}^{'}(f_{i}(v_{i}^{'})),\text{ } i=1,2,3,4,
\end{equation} }

\noindent where $f_{i}$ represents the four stages of the backbone and $g_{i}$ represents the extra convolutional layers. We do not add any layers to the last stage, so $g_{4}$ and $g_{4}^{'}$ can be instantiated as an identity function. As in most contrastive learning works, we use several MLP based projection heads $h_{i}$ to map each view features to the space where contrastive loss is applied:

\begin{equation}
\setlength\abovedisplayskip{0cm}
\setlength\belowdisplayskip{0cm}
\label{projection}
z_{i} = h_{i}(e_{i});\text{ } z_{i}^{’} = h_{i}(e_{i}^{’}).
\end{equation}

Thus, the overall loss function is fomulated as follows where $L_{contrast}$ could be any of the contrastive losses:

{\setlength\abovedisplayskip{0cm}
\setlength\belowdisplayskip{0cm}
\begin{equation}
L_{i} = L_{contrast}(z_{i}, z_{i}^{’}), \text{ } \text{ } \text{ } \text{ } L_{overall} = \sum\limits_{i=1}^{4} L_{i}.
\end{equation} }

In this way, the invariances of augmentation $t_{0}$ and $t_{1}$ are most widely distributed, while the invariance of $t_{4}$ is restricted to the deepest stage of the encoder.

In this paper, we instantiate $t_{0}$ as random cropping and resizing since ImageNet images are of different sizes, and cropping is considered the most fundamental augmentation for contrastive learning. $t_{1} \text{ } \sim t_{4}$ are selected without repetition in the following four augmentations as the classical pipeline: horizontal flipping, color jittering, converting to grayscale, and Gaussian blurring. Consistent with our intuition and motivation, the empirical study finds that instantiating $t_{i}$ according to the importance of each augmentation brings the best results, as shown in Section ~\ref{why_hier}.

\begin{table*}
\footnotesize
\renewcommand{\arraystretch}{1}
  \centering
  \begin{tabular}{@{}cccccccccccc@{}}
    \toprule
     \multirow{2}{*}{Model} & \multirow{2}{*}{Method} & \multicolumn{2}{c}{ImageNet} & \multicolumn{2}{c}{CUB-200} & \multicolumn{2}{c}{Flower-102} & \multicolumn{2}{c}{iNat-2019} & \multicolumn{2}{c}{Car-196} \\
     &  & Top-1 & Top-5 & Top-1 & Top-5 & Top-1 & Top-5 & Top-1 & Top-5 & Top-1 & Top-5 \\
    \midrule
    \multirow{2}{*}{ResNet50} & SimSiam~\cite{He2020MomentumCF} & 69.9 & 89.3 & 38.8 & 68.2 & 89.9 & 97.4 & 32.1 & 58.4 & 50.5 & 76.7 \\ 
     & SimSiam + Ours & \textbf{70.1} & \textbf{89.5} & \textbf{42.2} & \textbf{72.0} & \textbf{92.3} & \textbf{98.4} & \textbf{38.1} & \textbf{65.3} & \textbf{51.9} & \textbf{77.0} \\
    \midrule
    \multirow{2}{*}{ResNet50} & BT~\cite{Zbontar2021BarlowTS} & 67.0 & 87.6 & 33.8 & 62.9 & 89.1 & 97.2 & 31.9 & 58.1 & 37.2 & 64.3 \\
     & BT + Ours & \textbf{67.1} & \textbf{87.9} & \textbf{35.8} & \textbf{65.1} & \textbf{91.2} & \textbf{97.8} & \textbf{36.0} & \textbf{63.6} & \textbf{38.4} & \textbf{64.7} \\
    \bottomrule
  \end{tabular}
  \caption{Top-1 and top-5 accuracies (\%) under linear evaluation. All models use a ResNet-50 encoder pre-trained on ImageNet.}
  \label{main_results_1}
\end{table*}

\begin{table*}
\renewcommand{\arraystretch}{1}
\footnotesize
\renewcommand{\arraystretch}{1}
  \centering
  \begin{tabular}{@{}cccccccccccc@{}}
    \toprule
     \multirow{2}{*}{Model} & \multirow{2}{*}{Method} & \multicolumn{2}{c}{ImageNet-100} & \multicolumn{2}{c}{CUB-200} & \multicolumn{2}{c}{Flower-102} & \multicolumn{2}{c}{iNat-2019} & \multicolumn{2}{c}{Car-196} \\
    & & Top-1 & Top-5 & Top-1 & Top-5 & Top-1 & Top-5 & Top-1 & Top-5  & Top-1 & Top-5 \\
    \midrule
    \multirow{2}{*}{ResNet34} & SimSiam~\cite{He2020MomentumCF} & 63.9 & 88.7 & 29.1 & 58.5 & 76.1 & 91.9 & 16.9 & 36.6 & 19.6 & 42.0 \\
      & SimSiam + Ours & \textbf{67.1} & \textbf{90.5} & \textbf{33.9} & \textbf{64.2} & \textbf{83.6} & \textbf{94.5} & \textbf{23.2} & \textbf{44.5} & \textbf{20.7} & \textbf{42.7} \\
    \midrule
    \multirow{2}{*}{ResNet34} & BYOL~\cite{Grill2020BootstrapYO} & 66.5 & 89.1 & 30.0 & 59.4 & 76.6 & 92.3 & 17.8 & 37.6 & 20.5 & 42.6\\
      & BYOL + Ours & \textbf{69.3} & \textbf{91.3} & \textbf{33.6} & \textbf{64.0} & \textbf{83.2} & \textbf{94.1} & \textbf{24.4} & \textbf{47.5} & \textbf{21.9} & \textbf{43.8} \\
    \midrule
    \multirow{2}{*}{ResNet34} & BT~\cite{Zbontar2021BarlowTS} & 67.3 & 89.8 & 27.5 & 55.1 & 76.4 & 91.7 & 17.1 & 37.2 & 15.5 & 34.1 \\
      & BT + Ours & \textbf{69.1} & \textbf{91.7} & \textbf{29.4} & \textbf{58.5} & \textbf{82.1} & \textbf{94.9} & \textbf{23.3} & \textbf{45.0} & \textbf{17.0} & \textbf{36.2} \\
    \bottomrule
  \end{tabular}
  \caption{Top-1 and top-5 accuracies (\%) under linear evaluation. All models use a ResNet-34 encoder pre-trained on ImageNet-100.}
  \label{main_results_2}
\end{table*}

\subsection{Feature expansion with augmentation embeddings}
\label{expansion}
Some previous works take self-supervised tasks as auxiliary training in order to boost the performance of supervised learning~\cite{Lee2020SelfsupervisedLA}. These works utilize “label augmentation” to remove the unnecessary invariance brought by the self-supervised tasks. They augment the class label $y$ with the augmentation parameters used for the corresponding image. More specifically, the classifier should jointly predict the original label and the self-supervised label defined by the augmentation parameters (e.g., the rotation degree of $0^{\circ}, 90^{\circ}, 180^{\circ} \text{ } {\rm and} \text{ } 270^{\circ}$). 

There is no class label in self-supervised learning, and the augmentation parameters are contiguous float numbers in most cases. Thus, we propose to regard each view as the “label” of the other view and embed the augmentation parameters by a small network. In this way, we can expand each view feature with its augmentation embedding to remove the unnecessary invariance properly.

In practice, we utilize a linear layer $f_{aug}$ to embed the parameters of each augmentation to a vector. Taking color jittering as an example, we first get the four factors of brightness, contrast, saturation, and hue as $\left[b,c,s,h\right]$. We concatenate these scalars and embed them using a linear layer with the input size of 4: $e_{aug} = f_{aug}(\left[b,c,s,h\right])$. Then we concatenate the view feature and the augmentation embedding before feeding them to the projection head. So the Equation (\ref{projection}) is changed into:

\begin{equation}
z_{i} = h_{i}([e_{i}, e_{aug,i}]); \text{ } z_{i}^{’} = h_{i}([e_{i}^{’}, e_{aug,i}^{’}]).
\end{equation}

Previous works conjecture an intuitive understanding of the projection head: induced by the contrastive loss, the projection head removes information that may be useful for the downstream task, such as the color of objects~\cite{Chen2020ASF}. Thus, more useful information can be maintained in the backbone $f$. With the proposed feature expansion procedure, the projection head should consider both the feature and color-related information to meet the contrastive learning objective. In other words, we explicitly facilitate the projection head to do the job of removing color-related information. Therefore, useful information is more appropriately stored in the backbone $f$.

This paper also attempts to embed cropping parameters and combine the embeddings with color augmentation embeddings. For random cropping, we get four parameters representing the cropping box's location, height, and width $\left[x,y,h,w\right]$ and encode them in the same way as above. The detailed comparison of different augmentation embeddings is shown in Section~\ref{ablation}.

\section{Experiments}

\subsection{Experiment settings}
\label{settings}
\textbf{Datasets.} Most of our study for unsupervised pretraining is done using the ImageNet ILSVRC-2012 dataset~\cite{Deng2009ImageNetAL}. Additional pretraining experiments for ablation are done using a subset of ImageNet, 100-category ImageNet (IN-100). The split of this subset follows CMC~\cite{Tian2020ContrastiveMC}, which contains about 125K  images of 100 classes. The linear evaluation for the classification task is conducted on the ImageNet and some other datasets to measure the transfer capabilities as follows. The Caltech-UCSD Birds 2011 (CUB-200) dataset~\cite{Wah2011TheCB}, a fine-grained classification dataset of 200 bird species. VGG Flowers (Flowers-102)~\cite{Nilsback2008AutomatedFC}, a fine-grained dataset of 102 flower categories. The iNaturelist 2019 dataset (iNat-2019)~\cite{Horn2018TheIS}, a large-scale dataset with 268,243 images, which contains 1010 species of natural plants and animals. The Stanford Cars dataset (Car-196)~\cite{Krause20133DOR}, a fine-grained dataset which contains 16,185 images of 196 classes of cars. We measure the transfer capabilities to other tasks on two famous benchmarks, including VOC 2007~\cite{Everingham2009ThePV} for object detection and COCO~\cite{Lin2014MicrosoftCC} for both object detection and instance segmentation.

\textbf{Implementation details.} To learn the hierarchical augmentation invariance, we divide the backbone into four stages and extract the feature maps before each downsampling layer. The data augmentation $t_{1} \text{ } \sim t_{4}$ in the main experiments are color jittering, converting to grayscale, blurring, and horizontal flipping, respectively. The chosen strategy of these augmentation types is discussed in Section~\ref{why_hier}. To produce view features in different stages, we add several extra convolutional layers in shallow stages (the number of conv layers is 3,2,1 for the first three stages, respectively), which are discarded in the inference period.

The MLP-based projection head has the same structure as many previous works, which has three fully-connected layers. Note that we also add three projection heads to project the view features in shallow stages. 

To embed the augmentation parameters, we use a linear layer followed by BN and ReLU to project the parameters into a 512-dim vector for ImageNet pre-training experiments. This small network is also discarded during inference. In the main experiments, we expand the view features with the augmentation embedding in all stages that contain the corresponding augmentation invariance.

\textbf{Unsupervised pre-training.} We pre-train the ResNet-50 backbone on the 1000-category ImageNet training set following the classical protocol. We also pre-train the ResNet-34 backbone on 100-category ImageNet for ablation study using the same hyperparameter settings. Specifically, following SimSiam, we pre-train all the models for 200 epochs with the SGD optimizer. We use a learning rate of $lr$ $\times$ BatchSize $/$ 256 (linear scaling~\cite{Goyal2017AccurateLM}), with a base $lr=0.05$. The learning rate has a cosine decay schedule~\cite{Loshchilov2017SGDRSG,Chen2020ASF}. The weight decay is $0.0001$ and the SGD momentum is 0.9. The batch size is 256. For the sake of fair and direct comparison, we use the same settings for each baseline and our method, and only apply the three improvements to the baselines mentioned above.

\textbf{Linear evaluation, detection and segmentation.} We follow the linear evaluation protocol~\cite{Goyal2019ScalingAB}, in which the pre-trained model is fixed and only the additional linear classification layer is fine-tuned. For detection and segmentation tasks, we fine-tune the pre-trained models end-to-end in the target datasets. A Faster R-CNN detector~\cite{Ren2015FasterRT} is used for VOC and a Mask R-CNN detector~\cite{He2020MaskR} is used for COCO both with the ResNet50-C4 backbone implemented in Detectron2~\cite{Detectron2}. More details can be found in the appendix.

\subsection{Results}
\label{results}
First, we show the results of the linear evaluation in classification tasks where the model is pre-trained on ImageNet. In Table~\ref{main_results_1}, we compare our methods with corresponding baselines using the backbone of ResNet-50 pre-trained on ImageNet-1000. The top-1 and top-5 accuracies on several benchmarks are reported. Our methods significantly improve the performance on various classification downstream tasks, especially on fine-grained datasets, possibly because the representations from fine-grained datasets are more sensitive to data augmentations. For example, color information is more crucial to distinguish between flowers in Flower-102 and natural species in iNaturalist-2019. We also compare our methods with more baselines and different model architectures whose results are shown in Table~\ref{main_results_2}. Here we use the backbone of ResNet-34 pre-trained on ImageNet-100. The proposed method still leads to significant accuracy boost compared with the baseline models, reflecting the stable and universal effect of our method.

Next, we evaluate our representations for the localization based tasks of object detection and instance segmentation. For VOC 07 detection, we fine-tune on trainval2007 and report results on test2007 using the standard ${\rm AP_{50}, AP, AP_{75}}$ metric. For COCO detection and segmentation, we fine-tune on COCO 2017 train and report results on COCO 2017 val. Table~\ref{detection_results} shows that our method boosts the accuracies on both tasks, suggesting that our method improves the generalizability of the representations beyond classification tasks. 

\begin{table}
\setlength\tabcolsep{2pt}
\scriptsize
  \centering
  \begin{tabular}{@{}lccccccccc@{}}
    \toprule
    \multirow{2}{*}{Method} & \multicolumn{3}{c}{VOC 07 det} & \multicolumn{3}{c}{COCO det} & \multicolumn{3}{c}{COCO instance seg} \\
     & $\rm AP_{50}$ & $\rm AP$ & $\rm AP_{75}$ & $\rm AP_{50}$ & $\rm AP$ & $\rm AP_{75}$ & $\rm AP_{50}^{mk}$ & $\rm AP^{mk}$ & $\rm AP_{75}^{mk}$  \\
    \midrule
    SimSiam~\cite{He2020MomentumCF} & \textbf{72.3} & 45.8 & 49.4 & 55.0 & 36.0 & 38.6 & 51.8 & 31.9 & 33.9 \\
    SimSiam + Ours & 72.0 & \textbf{46.6} & \textbf{50.9} & \textbf{56.9} & \textbf{37.6} & \textbf{40.7} & \textbf{53.5} & \textbf{33.1} & \textbf{35.2} \\
    \midrule
    BT~\cite{Zbontar2021BarlowTS} & 71.2 & 45.6 & 49.4 & 59.6 & 40.0 & 43.0 & 56.1 & 35.1 & 37.2 \\
    BT + Ours & \textbf{71.7} & \textbf{46.0} & \textbf{49.7} & \textbf{60.0} & \textbf{40.2} & \textbf{43.5} & \textbf{56.7} & \textbf{35.1} & \textbf{37.6}\\
    \bottomrule
  \end{tabular}
  \caption{Transfer learning results on detection and segmentation. All models use a ResNet-50 encoder pre-trained on ImageNet.}
  \label{detection_results}
\end{table}

Above, we report the representation learning results using the exact composition of data augmentations as in classic contrastive learning methods. Next, we show that by restricting the impact scope of data augmentation, adding a specific type of data augmentation that was not used in the classical augmentation pipeline (e.g., rotation) can simultaneously improve the performance on both large-scale datasets and fine-grained datasets. We assume that classic contrastive learning methods introducing rotation data augmentation will lead to severe performance degradation because they treat each data augmentation equally. Instead, we restrict rotation invariances to the deeper layers of the model without weakening its augmentation strength. Following SimCLR~\cite{Chen2020ASF}, the rotation augmentation randomly rotates the image with the probability of 0.5 with one of \{0°, 90°, 180°, 270°\}. Table~\ref{rotation} shows the results of our methods compared with SimSiam baseline with and without rotation augmentation. First, by restricting rotation invariance to the deepest stage, our method outperforms the SimSiam~\cite{He2020MomentumCF} baseline with or without rotation augmentation. Second, compared with our method without rotation, adding rotation augmentation improves the performance on fine-grained datasets such as CUB-200 and Flower-102. Besides, adding rotation augmentation in our method brings a significantly smaller negative impact on the universal dataset (e.g., ImageNet) than the baseline. We also compare the results of adding rotation invariance starting from different stages of the model, which are shown in the last three lines in Table~\ref{rotation}. For our method, rotation-i means that the rotation invariance is added to the model from the ${\rm i^{th}}$ stage. These results indicate that enlarging the distribution of rotation invariance in the model is harmful to representation learning, which is in line with our intuition and motivation.

\begin{table}
\footnotesize
\setlength\tabcolsep{3pt}
  \centering
  \begin{tabular}{@{}lccc@{}}
    \toprule
    Method  & ImageNet-100 & CUB-200 & Flower-102 \\
    \midrule
    SimSiam~\cite{He2020MomentumCF} & 63.9 & 29.1 & 76.1 \\
    SimSiam + Ours & \textbf{67.1} & 33.9 & 83.6 \\
    \midrule
    SimSiam~\cite{He2020MomentumCF} + rotation & 61.3 & 28.4 & 76.4 \\
    Ours + rotation-4 & 66.2 & \textbf{34.1} & \textbf{84.7} \\
    \midrule
    Ours + rotation-1 & 60.9 & 31.8 & 84.2 \\
    Ours + rotation-2 & 62.0 & 32.0 & 84.7 \\
    Ours + rotation-3 & 64.3 & 32.9 & 84.0 \\
    \bottomrule
  \end{tabular}
  \caption{The impact on accuracy (\%) of adding random rotation to the augmentation pipeline. Rotation-i $({\rm i}=1,2,3,4)$ means the rotation invariance is added to the model from the ${\rm i^{th}}$ stage.}
  \label{rotation}
\end{table}

\begin{table}
\footnotesize
  \centering
  \begin{tabular}{@{}lcccc@{}}
    \toprule
    \multirow{2}{*}{Arrangement}  & \multicolumn{2}{c}{ImageNet-100} & \multicolumn{2}{c}{CUB-200} \\
      & Top-1 & Top-5 & Top-1 & Top-5 \\
    \midrule
    ${\rm \left[G,B,F,C\right]}$ & 64.5 & 88.0 & 32.9 & 63.6 \\
    ${\rm \left[B,F,C,G\right]}$ & 65.4 & 89.3 & 33.5 & 64.0 \\
    ${\rm \left[F,C,G,B\right]}$ & 66.3 & 90.0 & 33.2 & 63.8  \\
    ${\rm \left[C,G,B,F\right]}$ & \textbf{67.1} & \textbf{90.5} & \textbf{33.9} & \textbf{64.2} \\
    \bottomrule
  \end{tabular}
  \caption{Linear evaluation accuracy (\%) under different arrangement modes of augmentation types. (C: color jittering; G: converting to grayscale; B: blurring; F: flipping)}
  \label{stage}
\end{table}

\section{Discussion}
\label{discussion}
In this section, we analyze our methods in detail and explain why each of our improvement is effective.

\subsection{Why hierarchical augmentation works?}
\label{why_hier}
We assume that data augmentations have different degrees of both positive and negative effects on downstream tasks, which should be treated differently during training. So we propose to make the fundamental augmentation invariances more widely distributed in the encoder, and restrict some insignificant invariances to the deeper layers, namely learning hierarchical augmentation invariances. To verify this assumption, we change the chosen of 
$t_{1} \sim t_{4}$ in Equation~\ref{transform} and compare the results of linear evaluation on ImageNet and CUB-200. Table~\ref{stage} shows the results of four arrangement modes of augmentations, where C/G/B/F represent color jittering, converting to grayscale, blurring, and flipping, respectively. The results reveal that instantiating $t_{1}  \sim t_{4}$ according to the importance of each augmentation brings the best results, which confirms our assumption.

Besides, we compare our method with another two baselines, namely “Uniform” and “Hierarchical strength” in Table~\ref{aug_type}. For “Uniform”, we also apply multiple contrastive losses at different depths of the model. However, the augmentation invariances are uniform in different stages,which means that $T_{1} \sim T_{4}$ have the same composition of $t_{1} \sim t_{4}$ augmentations in Equation~\ref{transform}. For “Hierarchical strength”, we make all types of invariances distributed in all model stages, but the augmentation strength is stronger in deeper stages than the shallower stages. The training details are contained in the appendix. The results show that our method outperforms the two baselines. It confirms that only applying multiple objectives is not enough, and the add-one strategy is effective. Moreover, simply weakening the augmentation strength at shallow stages does not work since contrastive learning benefits from strong data augmentations.

\begin{table}
\small
  \centering
  \begin{tabular}{@{}lcccc@{}}
    \toprule
    \multirow{2}{*}{Method}  & \multicolumn{2}{c}{ImageNet-100} & \multicolumn{2}{c}{CUB-200} \\
     & Top-1 & Top-5 & Top-1 & Top-5 \\
    \midrule
    Uniform & 64.0 & 88.8 & 29.3 & 58.9 \\
    Hierarchical strength & 63.6 & 88.2 & 27.1 & 56.4 \\
    Hierarchical type (ours) & \textbf{67.1} & \textbf{90.5} & \textbf{33.9} & \textbf{64.2} \\
    \bottomrule
  \end{tabular}
  \caption{Comparison of accuracy (\%) with two designed baselines.}
  \label{aug_type}
\end{table}

\begin{table}
\small
\renewcommand{\arraystretch}{1}
  \centering
  \begin{tabular}{@{}cccc@{}}
    \toprule
    Task   &  Representation  &  Aug embedding & Accuracy (\%) \\
    \midrule
    \multirow{3}{*}{Color} & $e$ & $\times$ & 25.0   \\
                           & $h(e)$ & \checkmark & 12.1 \\
                           & $e$ & \checkmark   & \textbf{65.3} \\
    \bottomrule
  \end{tabular}
  \caption{Accuracy (\%) on the pretext task of predicting the transformation applied during the training using different representations. The accuracy of random guess is 10\%.}
  \label{aug_strength}
\end{table}

\vspace{2mm}
\subsection{Why feature extension with augmentation embeddings works?}
\label{why_expand}

Previous work~\cite{Chen2020ASF} conjectures that the projection head $h$ can implicitly remove information that may be useful for the downstream tasks. So the feature $e$ before the projection head $h$ in Equation~\ref{projection} maintains more useful information and is a better representation for downstream tasks. We expand each view feature $e$ with its augmentation embeddings. This modification explicitly promotes the projection head $h$ to remove useful augmentation-related information induced by the contrastive loss. So more information can be formed and maintained in $e$. To verify this hypothesis, following SimCLR~\cite{Chen2020ASF}, we conduct experiments that use either $h(e)$ or $e$ trained with or without feature expansion to learn to predict the transformation applied during training. Specifically, we evenly divide the color jittering augmentation into ten categories according to its strength. Then we take the view features and predict the relative distance of augmentation strength applied to them, which is a more difficult task than the original task in SimCLR~\cite{Chen2020ASF}. Table~\ref{aug_strength} shows that $e$ trained with feature expansion contains much more information about the transformation applied, while the other representations lose the information.

\subsection{Ablation study}
\label{ablation}

In this section, we conduct an ablation study on classification tasks with the model pre-trained on ImageNet-100. As random cropping and color jittering are fundamental augmentations for contrastive learning, we also attempt to embed cropping parameters and combine them with color augmentation embeddings as introduced in Section~\ref{expansion}. Table~\ref{tab:tb1} shows that embedding cropping parameters also leads to an accuracy boost compared with baseline but is less effective than embedding color parameters. Combining cropping embeddings with color embeddings has an unstable effect on different benchmarks. This is probably because we simply concatenate the two types of embeddings channel-wise during training, which could be improved in future work. The ablation study results shown in Table~\ref{tab:tb2} further demonstrate the effectiveness of the two proposed modules in our method, where “Hier. aug.” represents hierarchical augmentation invariance introduced in Section~\ref{hier} and “Aug. emb.” represents the feature expansion with augmentation embeddings introduced in Section~\ref{expansion}.

\begin{table}
\footnotesize
\setlength\tabcolsep{3pt}
\renewcommand{\arraystretch}{0.9}
\begin{floatrow}
\capbtabbox{
 \centering
 \begin{tabular}{cccc}
 \toprule
 \multicolumn{2}{c}{Embedding} & \multicolumn{2}{c}{Dataset} \\
\cmidrule(lr){1-2} \cmidrule(lr){3-4}
 Color & Crop & IN-100 & CUB \\
 \midrule
  $\times$ & $\times$ & 66.2 & 31.9 \\
  \checkmark & $\times$ & \textbf{67.1} & 33.9 \\
  $\times$ & \checkmark & 66.9 & 32.9 \\
  \checkmark & \checkmark & 66.7 & \textbf{34.0}\\
 \bottomrule
 \end{tabular}
}{
 \caption{Ablation study on embedding types (acc. /\%).}
 \label{tab:tb1}
}
\capbtabbox{
\centering
 \begin{tabular}{ccc}
 \toprule
 Hier. aug. & Aug. emb. & Acc. \\
 \midrule
  $\times$ & $\times$ & 63.9 \\
  $\times$ & \checkmark & 66.4 \\
  \checkmark & $\times$ & 66.2 \\
  \checkmark & \checkmark & \textbf{67.1} \\
 \bottomrule
 \end{tabular}
}{
 \caption{Ablation study on the two modules (acc. /\%).}
 \label{tab:tb2}
}
\end{floatrow}
\end{table}

\section{Conclusion}
\label{conclusion}

In this paper, we consider the drawbacks of the augmentation module in terms of type and strength. We first propose treating each augmentation differently and learning the hierarchical invariance in the backbone, making it more flexible in choosing augmentation types beforehand. We then propose expanding the view features with corresponding augmentation embeddings, which contribute to maintaining useful fine-grained information in view features. The proposed method can be combined with any contrastive learning method following the general framework. Experiments on several classification, detection and segmentation  downstream tasks demonstrate that the proposed method leads to a consistent and significant accuracy boost. We also conduct analytical and ablation experiments to explore the effectiveness of each component in our method. Some limitations of our method are included in the appendix.

{\small
\bibliographystyle{ieee_fullname}
\bibliography{egbib}
}

\end{document}